# Learning fuzzy clustering for SPECT/CT segmentation via convolutional neural networks

Junyu Chen[a)] and Ye Li
*Department of Electrical and Computer Engineering, Johns Hopkins University, Baltimore, MD, USA*
*Russell H. Morgan Department of Radiology and Radiological Science, Johns Hopkins Medical Institutes, Baltimore, MD, USA*

Licia P. Luna
*Russell H. Morgan Department of Radiology and Radiological Science, Johns Hopkins Medical Institutes, Baltimore, MD, USA*

Hyun W. Chung
*Department of Nuclear Medicine, Konkuk University Medical Center, Konkuk University School of Medicine, Seoul, South Korea*

Steven P. Rowe, Yong Du and Lilja B. Solnes
*Russell H. Morgan Department of Radiology and Radiological Science, Johns Hopkins Medical Institutes, Baltimore, MD, USA*

Eric C. Frey
*Department of Electrical and Computer Engineering, Johns Hopkins University, Baltimore, MD, USA*
*Russell H. Morgan Department of Radiology and Radiological Science, Johns Hopkins Medical Institutes, Baltimore, MD, USA*

**Purpose:** Quantitative bone single-photon emission computed tomography (QBSPECT) has the potential to provide a better quantitative assessment of bone metastasis than planar bone scintigraphy due to its ability to better quantify activity in overlapping structures. An important element of assessing the response of bone metastasis is accurate image segmentation. However, limited by the properties of QBSPECT images, the segmentation of anatomical regions-of-interests (ROIs) still relies heavily on the manual delineation by experts. This work proposes a fast and robust automated segmentation method for partitioning a QBSPECT image into lesion, bone, and background.

**Methods:** We present a new unsupervised segmentation loss function and its semi- and supervised variants for training a convolutional neural network (ConvNet). The loss functions were developed based on the objective function of the classical Fuzzy C-means (FCM) algorithm. The first proposed loss function can be computed within the input image itself without any ground truth labels, and is thus unsupervised; the proposed supervised loss function follows the traditional paradigm of the deep learning-based segmentation methods and leverages ground truth labels during training. The last loss function is a combination of the first and the second and includes a weighting parameter, which enables semi-supervised segmentation using deep learning neural network.

**Experiments and results:** We conducted a comprehensive study to compare our proposed methods with ConvNets trained using supervised, cross-entropy and Dice loss functions, and conventional clustering methods. The Dice similarity coefficient (DSC) and several other metrics were used as figures of merit as applied to the task of delineating lesion and bone in both simulated and clinical SPECT/CT images. We experimentally demonstrated that the proposed methods yielded good segmentation results on a clinical dataset even though the training was done using realistic simulated images. On simulated SPECT/CT, the proposed unsupervised model's accuracy was greater than the conventional clustering methods while reducing computation time by 200-fold. For the clinical QBSPECT/CT, the proposed semi-supervised ConvNet model, trained using simulated images, produced DSCs of 0.75 and 0.74 for lesion and bone segmentation in SPECT, and a DSC of 0.79 bone segmentation of CT images. These DSCs were larger than that for standard segmentation loss functions by $> 0.4$ for SPECT segmentation, and $> 0.07$ for CT segmentation with $P$-values $< 0.001$ from a paired t-test.

**Conclusions:** A ConvNet-based image segmentation method that uses novel loss functions was developed and evaluated. The method can operate in unsupervised, semi-supervised, or fully-supervised modes depending on the availability of annotated training data. The results demonstrated that the proposed method provides fast and robust lesion and bone segmentation for QBSPECT/CT. The method can potentially be applied to other medical image segmentation applications. © *2021 The Authors. Medical Physics published by Wiley Periodicals LLC on behalf of American Association of Physicists in Medicine.* [https://doi.org/10.1002/mp.14903]

## 1. INTRODUCTION

Prostate cancer is one of the most common cancers for men in the United States. In 2020, it was estimated that there would be 191,930 new cases of prostate cancer in the US.[1] Bone is a common site for metastasis in prostate cancer.[1] Although planar bone scintigraphy is often used to assess response and progression of bone lesions, quantitative single-photon emission computed tomography (SPECT) has been shown to provide improved quantification compared to planar imaging in other applications, and thus quantitative bone SPECT (QBSPECT) has the potential to provide more accurate estimates of prognostic metrics than planar imaging.[2-5] Image segmentation plays a vital role in the clinical application of QBSPECT, because quantifying tumor uptake and metabolic tumor burden requires accurate segmentation of bone and lesion structures. With the emergence of machine learning and deep learning, computer-assisted or computer-automated segmentation algorithms are poised to become a routine and essential part of medical image analysis in general. However, due to poor spatial resolution, noise, and contrast properties of SPECT images, developing automated segmentation methods for QBSPECT is challenging. While there has been substantial work on developing automated segmentation algorithms for other imaging modalities, manual delineation is still the most common method used clinically for SPECT imaging.[6,7] Manual segmentation is not only tedious and time-consuming, but also it can introduce intra- and inter-observer bias and variability. Thus, an automated algorithm that provides fast and accurate segmentation of anatomical or regions-of-interest (ROIs) for QBSPECT images is desired in both clinical practice and research.

Automated segmentation algorithms can be categorized into two groups. The first group is supervised methods, that is, those requiring a set of annotated input data. These methods can be highly effective, but they typically involve a training stage that requires a large number of images with ground truth labels. Deep learning is a representative of such methods, and it recently has become a major focus of attention in the image segmentation field due to its performance.[8] The vast majority of such work has been based on convolutional neural networks (ConvNets),[8-11] which usually requires a substantial number of accurately annotated training images. In nuclear medicine, various ConvNets-based studies have investigated their use for image segmentation. For instance, Liu et al.[12] designed a ConvNet for dopamine transporter brain SPECT segmentation that estimates the posterior mean of the fractional volume within each voxel. Built upon the idea of fusing multimodality information, Guo et al.,[13] and Li et al.[14] incorporated computed tomography (CT) images in training the ConvNet for improved lesion delineation in Positron emission tomography (PET). Despite their promising performance, ConvNets trained using a domain-specific dataset that lacks diversity compared to its target application often leads to poor generalizability on data from "unseen" domains.[15] Unsupervised methods represent another group. In contrast to supervised methods, they can be more robust to unseen domains, and are useful in the absence of substantial amounts of training data. Active contour models,[16] region growing, and clustering algorithms such as Gaussian mixture models (GMM),[17] or fuzzy C-means (FCM),[18] are the examples of unsupervised methods, where the segmentation is obtained based on characterizing intensity distributions of a given image. Much work has been devoted to developing unsupervised segmentation algorithms for nuclear medicine imaging.[19] For example, Abdoli et al.[20] presented a modified Chan-Vese active contour model[21] to take into account noise properties and heterogeneity lesion uptake observed in PET. With similar ideas, Layer et al.[22] and Belhassen et al.[23] incorporated spatial constraints into the objective functions of GMM and FCM to model the noise properties of PET and thus reduce misclassifications. However, these methods generally require computationally intensive optimization for each given image, and therefore they can be slow in practice.

This paper introduces a fast, accurate, and robust segmentation technique and applies it to QBSPECT and CT image segmentation, which takes advantage of both deep learning and intensity clustering. Specifically, we aim to develop a fully automated method for partitioning voxels into background, lesion, and bone in QBSPECT images. To this end, we propose a set of loss functions that are based on the classical FCM algorithm. An important property of the proposed loss functions is that they incorporate the fundamental idea of fuzzy clustering, where the amount (fuzziness) of the overlap between segmentation classes is controlled via a user-defined hyperparameter. The proposed model can be trained in three scenarios: in the first scenario, referred to as unsupervised or self-supervised, the model is trained by minimizing an unsupervised loss function (one that does not require labels) within an unlabeled training dataset, where the loss function depends only on the image intensity distributions; in the second scenario, which follows the traditional paradigm of deep learning, the model is trained by minimizing a supervised loss function that compares the output segmentation and the ground truth labels with the training dataset; in the final scenario, the model is trained in a semi-supervised manner by leveraging both the intensity distributions of the images and the available ground truth labels. The proposed models were implemented using a ConvNet that takes an $n$-D image volume and outputs a $C$-class probability vector for each voxel location. The models were trained purely on experimentally acquired images of a physical phantom and realistic, physics-based, simulated images of an anthropomorphic phantom. Nevertheless, they produced accurate segmentation results on the unseen patient images while offering substantial speed-up compared to the conventional clustering algorithms.

The paper is organized as follows. Section 2 introduces clustering-based (Section 2.A) and unsupervised/semi-supervised (Section 2.B) segmentation methods. The proposed methods are described in Section 2.C and the experimental setup in Section 2.D. Section 3 presents experimental results on both simulated and clinical data. Insights gained from the results are discussed in Section 4, and Section 5 presents the conclusion.

## 2. MATERIALS AND METHODS

### 2.A. Clustering-based Image Segmentation

Classical clustering-based segmentation methods, such as K-means,[24] FCM,[25] and GMM,[17] aim at characterizing the statistical properties of intensity levels in images by 'learning' the statistics of the intensity information from a given image.[26] These methods usually minimize an objective function to group together voxels that share similar intensity statistics. A widely used clustering method applied to medical image segmentation is FCM method,[25,27,28] due to its simplicity, robustness, and effectiveness. The objective function of the conventional FCM is written as follows:

$$J_{FCM} = \sum_{j \in \Omega} \sum_{k=1}^{C} u_{jk}^{q} \parallel y_i - v_k \parallel^2, \quad (1)$$

where $u_{jk}$ represents the membership functions for the $j^{th}$ voxel and $k^{th}$ class, $v_k$ is the class-centroid, $y_j$ is the observation (voxel value) at location $j$, $C$ indicates the number of classes, and $\Omega$ is the spatial domain of the image. The minimization of this objective function is given by:

$$\min_{u_{jk}, v_k} J_{FCM}, \ s.t. \begin{cases} \sum_{k=1}^{C} u_{jk} = 1, & \forall j \in \Omega \\ 0 < \sum_{j \in \Omega} u_{jk} < N, & k = 1, ..., C \end{cases} \quad (2)$$

Here $N$ is the total number of voxels in the image, and the parameter q in (1) is a weighting exponent that satisfies $q \geq 1$ and controls the amount of fuzzy overlap between clusters. Larger q values allow for a greater degree of overlap between the intensity levels in clusters (and vice versa). As $q \to 1$, the membership, $u_{jk}$, becomes crisper, and approaches a binary function. Note that when $q = 1$, the $J_{FCM}$ transforms to the well-known K-means problem.[18,28] Since FCM does not take spatial context information into consideration, the clustering result may be subject to noise and image artifacts. To overcome this issue, many efforts have been devoted to incorporating local spatial information into the FCM algorithm to improve the performance of image segmentation.[28-34] Ahmed et al.,[29] Pham et al.,[28] and Chuang et al.[33] introduced spatial constraints to the FCM's objective function to allow the labeling of a pixel to be influenced by the labels of its neighbors. Others[30,32,34] have incorporated spatial constraints and, at the same time, improving the computational speed. Among those, the Robust FCM (RFCM), proposed by Pham et al.,[28] is a straightforward but quite effective improvement on the original FCM. It incorporates a Markov-random-fields- (MRF)[35] based regularization term for penalizing changes in the value of the membership functions in local neighborhoods. The objective function of RFCM is described as follows:

$$J_{RFCM} = \sum_{j \in \Omega} \sum_{k=1}^{C} u_{jk}^{q} \parallel y_j - v_k \parallel^2 + J_{spatial}, \quad (3)$$

where

$$J_{spatial} = \beta \sum_{j \in \Omega} \sum_{k=1}^{C} u_{jk}^{q} \sum_{j \in N_j} \sum_{m \in M_k} u_{lm}^{q}, \quad (4)$$

In the above, term $N_j$ represents the neighboring voxels of voxel $j$, $M_k$ is a set containing $\{1, ..., C\} \backslash \{k\}$ (i.e., class numbers other than k), and $\beta$ controls the weight of spatial smoothness term. Using the Lagrange multiplier to enforce the constraint, and taking partial derivatives with respect to $v_k$ and $u_{jk}$, an iterative algorithm can be obtained:

$$u_{jk}^{n+1} = \frac{\left( \parallel y_j - v_i^n \parallel^2 + 2\beta \sum_{l \in N_j} \sum_{m \in M_k} \left( u_{lm}^n \right)^q \right)^{\frac{-1}{q-1}}}{\sum_{i=1}^{C} \left( \parallel y_j - v_i^n \parallel^2 + 2\beta \sum_{l \in N_j} \sum_{m \in M_i} \left( u_{lm}^n \right)^q \right)^{\frac{-1}{q-1}}}, \quad (5)$$

And

$$v_k^{n+1} = \frac{\sum_{j \in \Omega} \left( u_{jk}^{n+1} \right)^q y_j}{\sum_{j \in \Omega} \left( u_{jk}^{n+1} \right)^q}. \quad (6)$$

Since the objective function of RFCM operates on nothing but voxel-level information of the input image, this algorithm is thus unsupervised. However, the trade-off is the increment of computational complexity, as the objective function has to be iteratively minimized for each and every input image.

In order to improve computational efficiency and segmentation accuracy, some recent papers have presented self-supervised ConvNet-based approaches using modified objective functions of the classical algorithms as loss functions.[36,37] The following subsection introduces one such method that is closely related to the method proposed in this paper.

### 2.B. Unsupervised/semi-supervised ConvNet-based Image Segmentation

Recently, there have been a large number of supervised segmentation methods proposed that are based on deep neural networks. Among the different network architectures are fully convolutional neural networks (FCNs);[38] U-shape FCNs[10] such as[39,40] have achieved great success in biomedical image segmentation. These networks typically must be trained with a large number of training images. However, ground truth data is difficult to obtain in medical imaging. In order to compensate for a paucity of gold-standard images, weakly-supervised ConvNet-based approaches have been proposed.[41-43] These methods train neural networks with weakly annotated data such as bounding boxes for ROIs or using whole image-level labels. In Ref., [37] Kim et al. proposed a segmentation model that can be trained in a self-supervised manner using a novel loss function that is based on the Mumford-Shah functional.[44] This loss function enables the ConvNet to train itself using the unlabeled images as elements of the training data. This can be particularly useful for training ConvNet-based models in cases where gold-standard segmentation is not available. In a discrete setting, the loss function is formulated as follows:

$$\mathcal{L}_{MS} = \sum_{k=1}^{C} \sum_{j \in \Omega} z_{jk} \parallel y_j - c_k \parallel^2 + \lambda \sum_{k=1}^{C} \sum_{j \in \Omega} |\nabla z_{jk}|, \quad (7)$$

where $z_{jk}$ is the softmax output from a neural network that follows:

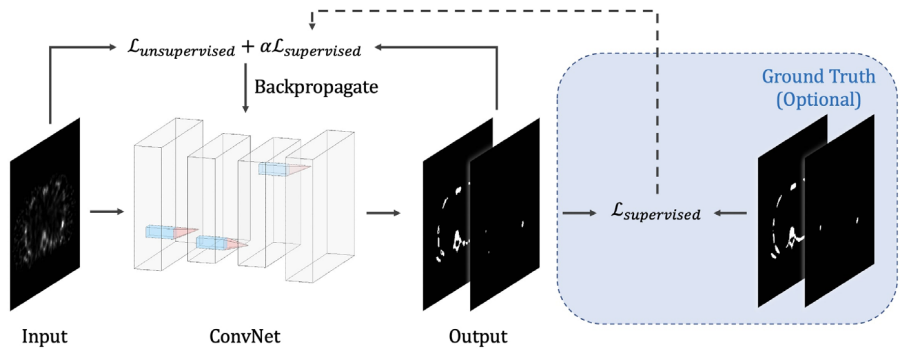

Fig. 1. Overview of the proposed method.

$$\sum_{k=1}^{c} z_{jk} = 1, \quad \forall j \in \Omega, \tag{8}$$

and $\left|\nabla z_{jk}\right|$ represents the Total variational (TV) norm of $z_{jk}$, where the grandient $\nabla(\cdot)$ can be approximated by the forward difference (i.e. $\nabla z_{jk} \approx z_{j+1k} - z_{jk}$). Finally, $c_k$ denotes the average voxel intensity value given by a similar form as (6):

$$c_k = \frac{\sum_{j \in \Omega} z_{jk} y_j}{\sum_{j \in \Omega} u_{jk}}. \tag{9}$$

Notice that if $\lambda$ is set to 0, then (7) is a special case of the FCM problem (1), in which $q = 1$. Then, the loss function can be thought of as a K-means clustering problem with TV regularization for suppressing noise in the membership function, $z_{jk}$. Kim et al. then proposed to incorporate a weighted supervised cross-entropy loss in addition to $\mathcal{L}_{MS}$ to consider semantic information from the ground truth labels[37]:

$$\mathcal{L}_{semi-MS}^{\alpha} = \mathcal{L}_{MS} + \alpha \mathcal{L}_{CE}. \tag{10}$$

Since this loss function consists of an unsupervised part that only requires information from the input image itself, and a supervised part that leverages ground truth segmentation, the resulting combined loss function is thus semi-supervised.

In nuclear medicine, image quality is severely affected by the partial volume effects (PVEs),[45] resulting in voxel values that are sums of the uptakes from or actually contain a mixture of different tissues. For this reason, algorithms like K-means or active contour that produce "hard" segmentation (i.e., a voxel can only belong to one class) may not be suitable for nuclear medicine. Here, we proposed incorporating the classical FCM objective function into ConvNet training owing to FCM's inherent suitability to nuclear medicine's low-resolution properties. This has the benefit that it takes advantage of the fast computation speed of ConvNets. The details of the proposed method are described in the following subsection.

## 2.C.  Proposed method

Let **y** be the input image defined over a spatial domain $\Omega \in \mathbb{R}^n$. For the rest of this paper, we focus on the 2-dimensional ($n = 2$) case; however, our implementation is dimension independent. We assume that the input image **y** is normalized in a preprocessing step.

Figure 1 shows an overview of the method. The ConvNet with parameters $\theta$, takes **y** as its input (i.e., $f(\mathbf{y};\theta)$), and outputs a $C$-channel probability map, where each channel corresponds to the probabilities of pixels in **y** belonging to a specific class. The optimal parameter $\hat{\theta}$ is determined by minimizing the expected loss function within a training dataset, i.e., $\hat{\theta} = argmin_{\theta} \mathcal{L}(\mathbf{y}; \theta)$. Here, we present three novel loss functions for the task of image segmentation. The first loss is proposed based on $J_{RFCM}$(3), which is denoted as $\mathcal{L}_{unsupervised}$ in Fig. 1. It is an unsupervised loss that depends on no ground truth labels but only on the input image **y**. The second loss, which is denoted as $\mathcal{L}_{supervised}$ in Fig. 1, is a modified version of $J_{FCM}$(1) that leverages the available ground truth information during training. Combining $\mathcal{L}_{unsupervised}$ and $\mathcal{L}_{supervised}$, we obtain the third loss function, which represents the semi-supervised application of the ConvNet model. A detailed description of the network architecture and the loss functions are discussed in the following sections.

### 2.C.1.  Network architecture

The neural network architecture, as seen from Fig. 2, is based on the recurrent convolutional neural network proposed by Liang et al. in Ref. [46]. The network consists of eight convolutional layers, where the first five are recurrent convolutional layers. For each recurrent convolutional layer, $T = 3$ time-steps where used, resulting in a feed-forward subnetwork with a depth of $T + 1 = 4$.[46] Each convolutional layer has a kernel size of 3 by 3, and it is followed by batch normalization and a Rectified linear unit (ReLU). In the final layer, a 3-channel soft-max activation function is used, where each channel represents the probability of being classified as background, bone, or lesion.

### 2.C.2.  Unsupervised loss function

We propose to model the membership functions, $u$, in the objective function of RFCM (3) by using the softmax output of the last layer of the ConvNet model, i.e. $f(\mathbf{y};\theta)$, using:

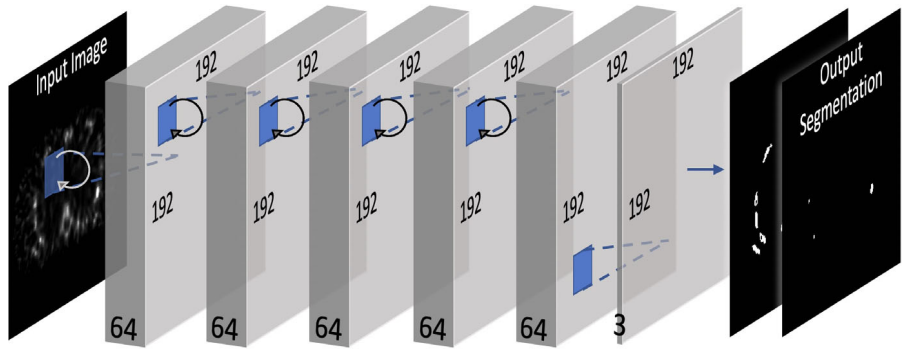

Fig. 2. ConvNet architecture.

$$\mathcal{L}_{RFCM}\ (\mathbf{y};\theta) = \sum_{j\in\Omega}\sum_{k=1}^{C} f_{jk}^{q}(\mathbf{y};\theta)\| y_j - v_k \|^2$$

$$+ \beta \sum_{j\in\Omega}\sum_{k=1}^{C} f_{jk}^{q}(\mathbf{y};\theta)\sum_{l\in N_j}\sum_{m\in M_k} f_{lm}^{q}(\mathbf{y};\theta), \qquad (11)$$

where $f_{jk}(\mathbf{y};\theta)$ is the $k^{th}$ channel softmax output from the ConvNet at location $j$, and the class mean $v_k$ shares a similar form as follows (6):

$$v_k = \frac{\sum_{j\in\Omega} f_{jk}^{q}(\mathbf{y};\theta) y_j}{\sum_{j\in\Omega} f_{jk}^{q}(\mathbf{y};\theta)}. \qquad (12)$$

This differentiable loss function relies solely on characterizing the intensity statistics of the input image $\mathbf{y}$ that are independent of ground truth labels. This property makes it an unsupervised loss function that can be minimized in a self-supervised manner using deep neural networks; that is, the loss function enables the training of a network using unlabeled images. As described in Ref. [37,47], the self-supervised training framework can be interpreted as an unfolded fixed number of iterations for solving for the membership functions (5). Next, we describe the techniques for incorporating a supervised loss function.

### 2.C.3. Semi-supervised/Supervised Loss Functions

While networks trained using unsupervised loss functions can be more robust to test images from an "unseen" domain, their performances are generally limited due to the sole dependence on pixel-level intensity information. Supervised losses, on the other hand, leverage the prior knowledge that includes knowledge of ground truth to learn about semantic and shape information. However, the trade-off is that these supervised ConvNet models may suffer from overfitting and lack of generalizability (to images differing substantially from those used in training) depending on the size of the training dataset.[48] In the semi-supervised setting described in Ref. [37], the ConvNet is trained with both supervised and unsupervised loss functions so that the network can take into account the intensity statistics in an individual image while embracing the supervised information provided by the ground truth. The proposed unsupervised loss function, $\mathcal{L}_{RFCM}$, can be paired with any supervised loss function in a

similar way as described in Ref. [36,37]. That is:

$$\mathcal{L}(\mathbf{y};\theta) = \mathcal{L}_{RFCM}((\mathbf{y};\theta)) + \alpha\mathcal{L}_{supervised}(\mathbf{y}, \mathbf{g};\theta), \qquad (13)$$

where $\alpha$ is a weighting parameter for controlling the strength of the supervised term, and $\mathbf{g}$ denotes ground truth label map. It is, however, difficult to use supervised loss functions such as Dice similarity (DSC) loss or cross-entropy (CE) because they are not immediately compatible with the 'fuzziness' of the classification produced by FCM and inherent in $\mathcal{L}_{RFCM}$, and that they may take different value ranges from $\mathcal{L}_{RFCM}$. Inspired by,[49] here we propose a novel supervised segmentation loss function on the basis of the FCM objective function:

$$J_{FCM_{label}} = \sum_{j\in\Omega}\sum_{k=1}^{C} u_{jk}^{q}\| g_{jk} - \mu_k \|^2, \qquad (14)$$

where the membership function, $u_{jk}$, can be modeled using $f_{jk}(\mathbf{y};\theta)$, which is the ConvNet's softmax output, $g_{jk}$ is the ground truth label at location $j$ for $k^{th}$ class, and $\mu_k$ is the class mean computed within $\mathbf{g}$. The ground truth images, $\mathbf{g}$, have $C$ channels, where each channel is a binary segmentation mask specifying the spatial domain of class $k\in\{1, ..., C\}$. The class mean $\mu_k$ can be simply defined in advance as a constant 1, i.e., $\mu_k = 1$ (by suitably rescaling $g_{jk}$). The loss function can thus be defined as follows:

$$\mathcal{L}_{FCM_{label}}(\mathbf{y};\theta) = \sum_{j\in\Omega}\sum_{k=1}^{C} f_{jk}^{q}(\mathbf{y};\theta)\| g_{jk} - 1 \|^2, \qquad (15)$$

This loss function incorporates the main idea of FCM, for which the parameter $q$ is included to regulate the fuzzy overlap between softmax channels. Thus, the proposed semi-supervised loss function is thus written as follows:

$$\mathcal{L}_{semi-RFCM}^{\alpha}(\mathbf{y};\theta)\ = \mathcal{L}_{RFCM}(\mathbf{y};\theta) + \alpha\mathcal{L}_{FCM_{label}}(\mathbf{y};\theta). \qquad (16)$$

### 2.D. Experimental setup

The proposed ConvNets were implemented and tested using TensorFlow.[50] Unless otherwise specified, all the experiments are carried out in Python 3.8 on a PC with AMD Ryzen 9 3900X 3.79 GHz CPU and an NVIDIA Titan RTX GPU (24 Gb). We investigated the performance of the proposed model on the task of bone and lesion segmentation in SPECT/CT images. We first (Section 3.1) present a hyper-parameter analysis for the proposed unsupervised loss on a dataset of realistic QBSPECT simulations. In the second experiment (Section 3.2), we present a series of segmentation experiments using the proposed supervised loss in which the performance is compared to several widely used supervised loss functions on the realistic QBSPECT simulations. In the third experiment (Section 3.3), we demonstrate the effectiveness of the proposed semi-supervised, simulation-trained ConvNet on segmenting lesion and bone from a clinical SPECT scan; the result from the ConvNet was compared to VOIs manually delineated by an experienced physician, which served as a gold-standard. Since CT images provide higher resolution and likely better anatomical information for bone structures, in the final experiment (Section 3.4), we applied and evaluated the proposed semi-supervised loss to the task of segmenting bones in a clinical CT scan.

### 2.D.1. Dataset

The proposed algorithm was assessed on three datasets, where the training and a testing set consist of the attenuation maps generated from the XCAT phantom[51,52] and their corresponding simulated SPECT images; another testing set is a set of clinical SPECT/CT bone scans obtained from an institutional-review-board-approved protocol. A brief description of the data is provided below:

*XCAT attenuation maps & SPECT simulations:* The bone SPECT simulations were generated based on the 3D activity and attenuation distributions obtained using a population of highly realistic NURBS-based XCAT phantoms. The phantom variations were generated using our previously proposed deformable-registration-based phantom generation method.[53,54] Nine XCAT variations were created by mapping a single XCAT attenuation map to capture the anatomical variations of the CT scans from nine patients. The values of the XCAT attenuation maps were calculated on the basis of the material compositions and the attenuation coefficients of the constituents at 140 keV, the photon energy of Tc-99m methylene diphosphonate (MDP) bone imaging. The activity distribution modeled that seen in Tc-99m. The uptake of bone was modeled as 8–13 times that of the soft-tissue background. We modeled sclerotic bone lesions in the simulation with increased attenuation coefficient and radiopharmaceutical uptake. The attenuation coefficient was varied randomly with uniform probability in the range 1.13–1.27 times the attenuation coefficient of cortical bone. The uptake in lesions had a mean of 4 times that of the bone, and varied randomly with a uniform distribution by a factor of 0.5–1.75 around that mean. SPECT projections were simulated using an analytic projection algorithm that realistically models attenuation, scatter assuming a 20% wide energy window centered at 140 keV, and the spatially-varying collimator-detector response (CDR) of a low-energy high-resolution collimator on a GE Discovery 670 scanner.[55,56] Projections were simulated at 256 projections over 360° and collapsed after projection to model an acquisition voxel size of 0.22 cm. The acquisition time was such that it modeled a whole-body scan time of 30 minutes, and the projections were scaled appropriately, and Poisson noise was simulated. Bone SPECT images were then reconstructed using a quantitative reconstruction method described in Ref. [5], which is based on the ordered subsets expectation-maximization algorithm[57] (OS-EM) and modeled attenuation, scatter, and the CDR. We used 2 iterations of 10 subsets per iteration during reconstruction. No post-filtering was performed. A total of nine 3D SPECT volumes were simulated using the XCAT phantom variations.

*Clinical patient SPECT/CT scans:* We used 11 clinical SPECT and 12 clinical CT images in this study. The SPECT scans were acquired on a Siemens Symbia T16 SPECT/CT system. The bone SPECT acquisitions were performed using our clinical bone SPECT protocol (120 views per bed position, and two energy windows with widths of 15% and centered at 140 KeV and 119 keV with the latter serving as a scatter window), followed by CT acquisition. The clinical CT images were acquired from 12 patients, with 8 on the Siemens SPECT/CT system and 4 on a GE CT system. Both SPECT and CT images were reconstructed using scanner software. The SPECT reconstructions were obtained using Flash 3D/OS-EM (with default numbers of iteration and subsets) with attenuation, energy-window-based scatter, and geometric collimator-detector response compensations enabled. Bone and lesion in the SPECT and CT images were manually delineated by consensus of two experienced radiologists, where the initial segmentation was produced by a radiology fellow and verified by an attending physician. Two different fellows produced the initial segmentation.

### 2.D.2. Baseline methods

We compared the results obtained using the proposed loss functions with those from two widely used supervised loss functions for ConvNet-based image segmentation, Dice loss ($\mathcal{L}_{DSC}$) and Cross-entropy loss ($\mathcal{L}_{CE}$). We also compared the results to a newly proposed unsupervised loss function, Mumford-Shah loss ($\mathcal{L}_{MS}$),[37] and a semi-supervised variant of it. For fair comparison, the same network architecture (described in section 2.3.1) was used for all loss functions. The baseline loss functions are briefly described below, and we refer **g** as the ground truth segmentation here:

*Dice loss:* Dice similarity coefficient (DSC) quantifies the overlap between two segmentation labels, its loss function is defined by:

$$\mathcal{L}_{DSC}(\mathbf{y};\theta) = 1 - \frac{2|f(\mathbf{y};\theta) \cap \mathbf{g}|}{|f(\mathbf{y};\theta)| + |\mathbf{g}|}. \tag{17}$$

*Cross-entropy loss:* CE is computed as the log loss, summed over all pixel locations and all the possible classes. It can be written as follows:

$$\mathcal{L}_{CE}(\mathbf{y};\theta) = -\frac{1}{\Omega}\sum_{j\in\Omega}\sum_{k=1}^{C}g_{jk}\log f_{jk}(\mathbf{y};\theta). \qquad (18)$$

*Mumford-shah loss:* The unsupervised and the semi-supervised Mumford-Shah losses[37] were previously given in (7) and (10).

*Fixed thresholding:* The proposed method was also compared to the fixed-threshold (FT) method. The threshold value was set to 42% of the maximum voxel value of the image for segmenting lesion, which has been proposed as a threshold value in SPECT imaging.[6,58] We used 2% of the maximum voxel value based on our empirical experiments for bone segmentation in SPECT. The intensity of bone in CT images, according to the literature, ranges from 400 to 800 in Hounsfield Unit (HU).[59-62] In this study, the threshold was empirically set to be 400 HU for identifying bone in CT.

### 2.D.3. Evaluation metrics

We used four commonly used metrics in image segmentation to quantify 3D segmentation performance. In addition to these common metrics, the recently proposed surface DSC was also used to take segmentation variability into consideration. The five evaluation metrics are briefly introduced below:

*Dice Similarity Coefficient (DSC):* DSC is a measure of the spatial overlap accuracy of the segmentation results.

*Recall:* Recall, also referred to as sensitivity, measures the completeness of the positive segmentation relative to the ground truth segmentation.

*Precision:* Precision effectively demonstrates the purity of the positive detection relative to the ground truth segmentation.

*Intersection over Union (IoU):* Similar to DSC, IoU is a metric for quantifying segmentation overlap between prediction and ground truth, it is also referred to as the Jaccard index.

*Surface dice similarity coefficient:* Inter- and intra-observer variability exists if the delineations were performed by different observers (i.e. physicians). The surface DSC[63] was proposed to take segmentation variability into consideration. In this study, the training dataset consisted only of simulated images with ground truth boundaries. Thus, the ConvNets did not learn how the humans would delineate the object-of-interest or the variability in delineation by different observers. Thus, the segmentation provided by the ConvNets might vary, and especially at the boundaries, from the physicians' delineation of the same object. The conventional DSC weighs all regions of misplaced segmentation equally, and it has been shown that the DSC has a strong bias toward larger objects (i.e. segmenting larger objects often yields higher DSC). However, bone metastases observed in the SPECT images can be small in size. Therefore, even small variations in the segmentation boundary can result in a low DSC, even when these variations are well within an observer's ability to accurately delineate a tumor. To overcome these shortcomings, Nikolov et al. in Ref. [63] proposed the surface DSC. The surface DSC provides a measure of the fraction of a region's contour that must be redrawn compared to the "gold-standard." The surface DSC considers a tolerance (in mm) within which the delineations' variations are clinically acceptable. A surface, $\mathcal{S}$, refers to the boundary of a segmentation mask, $\mathcal{M}$ (i.e., $\mathcal{S} = \partial\mathcal{M}$), and the area of the surface is defined as follows:

$$\mathcal{S} = \int_{\mathcal{S}} d\boldsymbol{\sigma}, \qquad (19)$$

where $\boldsymbol{\sigma}\in\mathcal{S}$ is a point on $\mathcal{S}$. The mapping (denote as $\xi$) from $\boldsymbol{\sigma}$ to a point, $\mathbf{x}$, in $\mathbb{R}^3$ is given as $\xi:\mathcal{S}\to\mathbb{R}^3$. Then the border region, $\mathcal{B}$, for the surface $\mathcal{S}$ at a distance tolerance $\tau$ is given by:

$$\mathcal{B}^{(\tau)} = \{\mathbf{x}\in\mathbb{R}^3 | \exists\boldsymbol{\sigma}\in\mathcal{S}, \parallel\mathbf{x}-\xi(\boldsymbol{\sigma})\parallel \leq\tau\}. \qquad (20)$$

Finally, the surface DSC at tolerance $\tau$ can be written as follows:

$$DSC_{surf} = \frac{|\mathcal{S}_p\cap\mathcal{B}_g^{(\tau)}| + |\mathcal{S}_g\cap\mathcal{B}_p^{(\tau)}|}{|\mathcal{S}_p| + |\mathcal{S}_g|}, \qquad (21)$$

where $p$ and $g$ denote, respectively, the prediction from the method under test and the "gold-standard". Similar to the convention DSC, $DSC_{surf}$ ranges from 0 to 1. A $DSC_{surf}$ of 0.95 simply means that 95% of the segmentation boundary was within $\tau$ mm of the "gold-standard" while 5% needs to be redrawn.[63] The tolerance $\tau$ was set to be one voxel width for lesion segmentation and twice the voxel width for bone segmentation. The tolerance used for these two objects was based on our observation of the variations in organ boundaries of the "gold-standard" delineations provided by our two physicians. We observed a greater deviation for the normal bone, possibly because segmenting them required a much larger amount of work, and possibly because they deemed accurate segmentation of the normal bones less important than the lesions.

Because the patient studies were acquired under different imaging protocols (i.e., varying from 1-bed to 3-bed positions), and considering the variations and imbalances in the number of lesions and lesion sizes in different patients, a weighted average scheme was used to assess the segmentation performances on the patient studies:

$$E[s] = \sum_{p_i}\frac{v_{p_i}}{\sum_i v_{p_i}}s_{p_i}, \qquad (22)$$

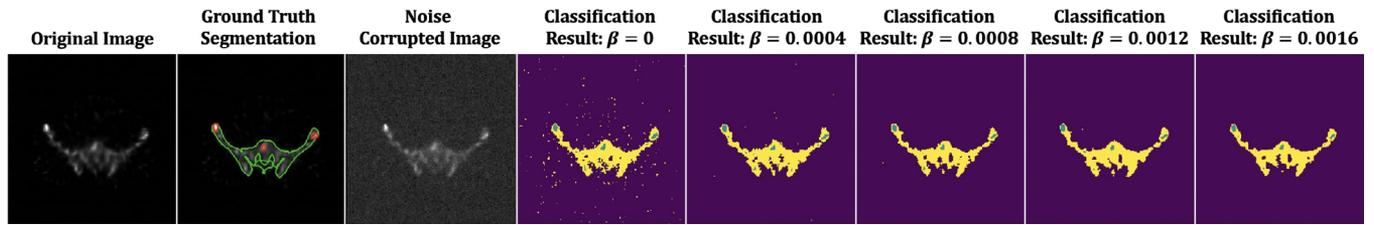

FIG. 3. The proposed unsupervised model applied to a 2D QBSPECT image. The first image: QBSPECT image. The second image: Ground truth segmentation of bone (green) and lesion (red). The third image: QBSPECT image corrupted by Gaussian noise with $\sigma = 0.01$. The fourth image to the last image: Maximum membership classification results using different $\beta$.

where $s \in \{DSC, Recall, Precision, IoU\}$ denotes the evaluation metric, and $v_{p_i}$ denotes the volume (based on the "gold-standard") of the object-of-interest (i.e., lesion or bone) in a patient $p_i$. Notice that the $DSC_{surf}$ scores were averaged across all patient studies because $DSC_{surf}$ is independent of the sizes of the object-of-interest.[63]

## 3. RESULTS

### 3.A. Unsupervised Loss on QBSPECT simulations

The ConvNet was trained using the proposed RFCM loss function in a self-supervised manner without the use of ground truth information. We used the realistic QBSPECT simulations for both training and testing. The effectiveness of the parameters $\beta$ and $q$ was first tested. We then compared the proposed model to the unsupervised Mumford-Shah loss,[37] and two clustering methods, FCM[18] and RFCM.[28] We first studied the effects of the two hyper-parameters in (11). The first parameter is $\beta$, for controlling the strength of the regularizer. We tested $\beta$ in a range of 0 to 0.0016 with a step size of 0.004. Figure 3 shows the qualitative results provided using different $\beta$ values. From left to right, the images are the original image, the original image with the addition of random Gaussian noise ($\sigma = 0.01$), classification results from five different $\beta$ values. As visible from the classification images, larger $\beta$ values lead to cleaner classification results due to the increased strength of spatial smoothness prior, which tends to suppress isolated pixel classification. Notice that the high noise level here is not clinically realistic, but it was used to show the effectiveness of the proposed spatial regularization. Thanks to the good intrinsic noise-suppressing ability of the ConvNet architecture,[64] even when beta was set to zero (i.e., without the spatial smoothness prior), the unsupervised ConvNet still yielded a robust segmentation result. The effectiveness of the spatial prior is also proven by the quantitative results shown in Table 1, where the segmentation accuracy gradually improves as $\beta$ increases. We evaluated the sensitivity of the resulting segmentation to the fuzzy exponent, $q$, in values of $q \in [1, 2, 4]$. Some qualitative results are shown in Fig. 4. The last four columns in the left panel of Fig. 4 exhibit the maximum membership classification (i.e., the class that has the maximum membership value) results and the membership functions for the three classes (background, bone, and lesion). When $q = 1$ (the first row), the membership functions are nearly binary, but as $q$ increases from 1 to 4, the degree of overlap between classes becomes higher, resulting in "fuzzier" membership functions. We also compared the proposed loss function to the unsupervised $\mathcal{L}_{MS}$.[37] As we previously described in section 2.2, $\mathcal{L}_{MS}$ has a similar form to the K-means problem. So its resulting softmax output, $z_{jk}$, is thus, almost binary, as shown in the first row of the left panel in Fig. 4. Finally, we compared the ability of noise regularization between the ConvNets trained by the different unsupervised loss functions. We used a noise-free dataset for training the ConvNet models, and then we used the Gaussian noise corrupted ($\sigma = 0.01$) images for testing. The parameter, $\lambda$, was set to be $= 10^{-9}$ for $\mathcal{L}_{MS}$ based on the empirical experiments, and $\beta = 0.0016$ and $q = 2$ for $\mathcal{L}_{RFCM}$. For the clustering algorithms, we set $q = 2$ for both FCM and RFCM, and $beta = 0.01$ for RFCM. Visually, as shown in the right panel of Fig. 4, the ConvNet trained using $\mathcal{L}_{RFCM}$ delineated the true objects in the image and was more robust to the noise than $\mathcal{L}_{MS}$. The FCM failed on the noisy images, resulting in unacceptably noisy segmentations. The RFCM, however, achieved comparable results to the segmentation obtained using ConvNet-based methods. Quantitatively, as shown in Table II, the performance decreased as the testing images were corrupted by noise for both unsupervised loss functions. However, the proposed $\mathcal{L}_{RFCM}$ provided higher values of the quantitative metrics than $\mathcal{L}_{MS}$ regardless of the noises. The best lesion segmentation score was achieved using the proposed unsupervised ConvNet, and the best bone segmentation score was achieved by the RFCM. However, as shown in Table III, compared to the RFCM, the proposed unsupervised ConvNet was $\frac{2.868}{0.013} \approx 221$ times faster.

TABLE I. Comparisons of various segmentation quality metrics for various values of the regularization parameter, $\beta$, for the proposed unsupervised loss.

| | Lesion | | | | Bone | | | |
|---|---|---|---|---|---|---|---|---|
| $\beta$ | DSC | Recall | Precision | IoU | DSC | Recall | Precision | IoU |
| 0 | 0.465 | 0.779 | 0.131 | 0.303 | 0.548 | 0.625 | 0.516 | 0.377 |
| 0.0004 | 0.472 | 0.775 | 0.134 | 0.309 | 0.562 | 0.608 | 0.564 | 0.391 |
| 0.0008 | 0.483 | 0.752 | 0.141 | 0.318 | 0.572 | 0.604 | 0.596 | 0.401 |
| 0.0012 | 0.500 | 0.716 | 0.154 | 0.333 | 0.586 | 0.600 | 0.636 | 0.414 |
| 0.0016 | 0.495 | 0.734 | 0.147 | 0.329 | 0.578 | 0.584 | 0.643 | 0.407 |

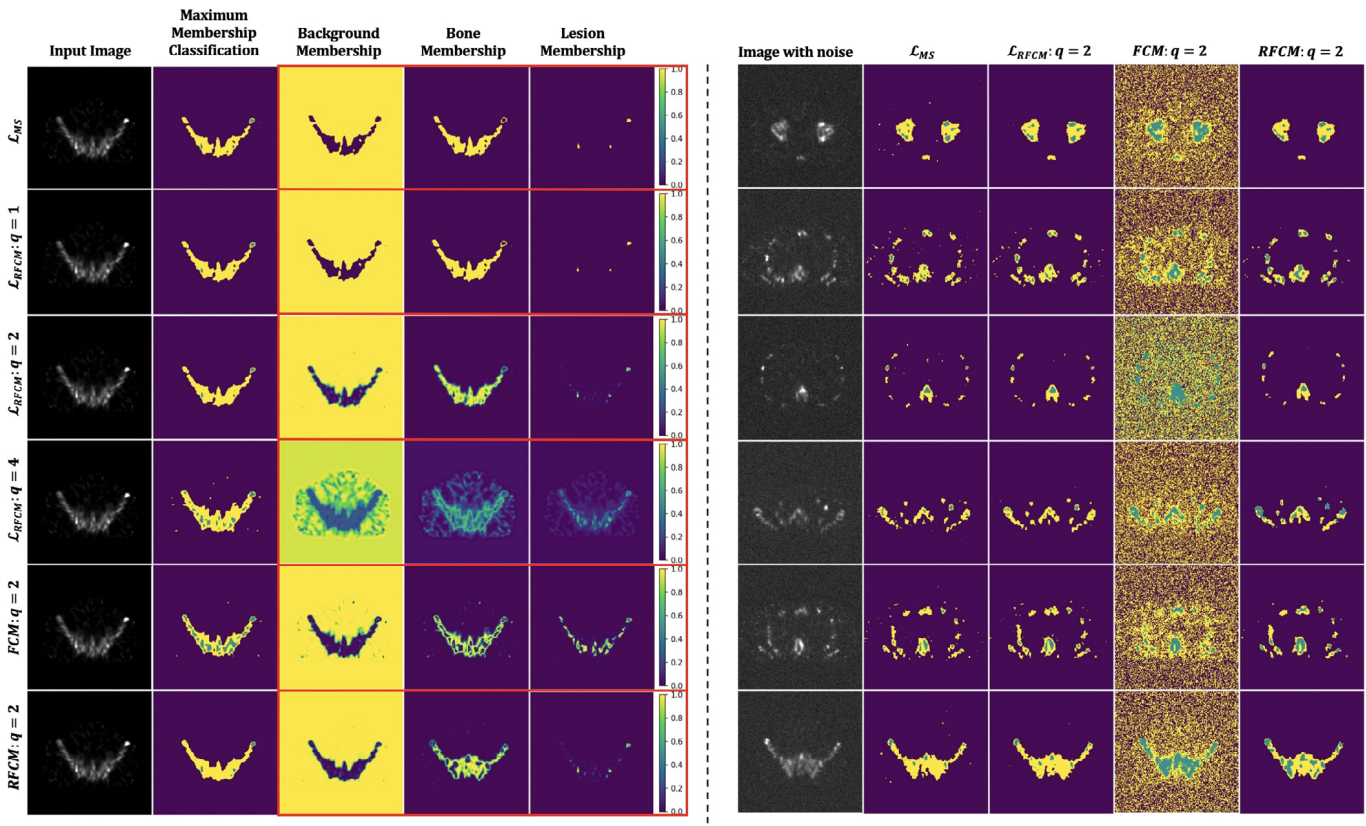

FIG. 4. Visual comparison of segmentation results obtained from the ConvNets trained using the proposed unsupervised loss function and other unsupervised segmentation methods. Left panel: Clustering results and membership images generated using $\mathcal{L}_{MS}$[37] (the first row), the proposed $\mathcal{L}_{RFCM}$ (the second through fourth rows), FCM[18] (the second through last rows), and RFCM[28] (the bottom row). Right panel: Comparisons between the maximum membership classification results generated using two unsupervised losses (the second and the third columns), FCM (the fourth column), and RFCM (the last column) on the Gaussian noise ($\sigma = 0.01$) corrupted images.

### 3.B. Supervised Loss on QBSPECT simulations

In this experiment, we used the ground truth labels to train the ConvNets using the supervised loss functions applied to the realistic simulated QBSPECT images. We first evaluated the effectiveness of the fuzzy exponent, $q$, in (13), with values of $q \in [1, 2, 4, 6]$. Then, we compared the proposed supervised loss, $\mathcal{L}_{FCM_{label}}$, with two widely used loss functions for ConvNet-based segmentation, $\mathcal{L}_{DSC}$ and $\mathcal{L}_{CE}$, on a set of simulated QBSPECT images. The proposed supervised loss, $\mathcal{L}_{FCM_{label}}$, based on the FCM formulation that incorporates the parameter $q$ for controlling the fuzzy overlap between channels of the softmax output. Figure 5 illustrates some examples of the resulting segmentation and membership functions (i.e., outputs of the last softmax layer) for $\mathcal{L}_{DSC}$, $\mathcal{L}_{CE}$, and $\mathcal{L}_{FCM_{label}}$ with different values of $q$. The third through last rows are the results produced using $\mathcal{L}_{FCM_{label}}$ with $q$ values ranging from 1 to 6. We observed the same trend as in Fig. 4, where larger $q$ values tended to produce fuzzier membership functions. Note that the ConvNet trained using $\mathcal{L}_{DSC}$ and $\mathcal{L}_{CE}$ produced hard (i.e., non-fuzzy) membership functions, which are similar to the results produced using $\mathcal{L}_{FCM_{label}}$ with $q = 1$. Table IV shows the quantitative evaluations for segmentation of the realistic QBSPECT simulations. Except for $q = 1$, the proposed $\mathcal{L}_{FCM_{label}}$ yielded comparable scores to $\mathcal{L}_{DSC}$ and $\mathcal{L}_{CE}$, while providing an interesting property that the resulting membership functions was not necessarily hard. This could potentially be beneficial when quantifying activity inside regions in SPECT images.

### 3.C. Semi-supervised Loss on Clinical SPECT

In this experiment, we used the proposed semi-supervised loss to train the ConvNet using the simulated QBSPECT images. Before training, data augmentation was applied to the training dataset, where a Gamma correction[65,66] was performed with $\gamma \in [0.9, 1.1]$. The training and testing image

TABLE II. Quantitative comparison results of unsupervised segmentation methods on the realistic QBSPECT simulations (top table) and the Gaussian noise corrupted simulation images (bottom table).

Noise-free Image

| | Lesion | | | | Bone | | | |
|---|---|---|---|---|---|---|---|---|
| | DSC | Recall | Precision | IoU | DSC | Recall | Precision | IoU |
| $\mathcal{L}_{MS}$[37] | 0.487 | 0.739 | 0.145 | 0.322 | 0.567 | 0.603 | 0.583 | 0.396 |
| $\mathcal{L}_{RFCM}$ | 0.500 | 0.716 | 0.154 | 0.333 | 0.586 | 0.600 | 0.636 | 0.414 |
| $FCM$[18] | 0.367 | 0.796 | 0.104 | 0.225 | 0.660 | 0.881 | 0.535 | 0.493 |
| $RFCM$[28] | 0.435 | 0.750 | 0.136 | 0.278 | 0.729 | 0.803 | 0.716 | 0.574 |

Image corrupted by Gaussian noise ($\sigma = 0.01$)

| | Lesion | | | | Bone | | | |
|---|---|---|---|---|---|---|---|---|
| | DSC | Recall | Precision | IoU | DSC | Recall | Precision | IoU |
| $\mathcal{L}_{MS}$[37] | 0.478 | 0.735 | 0.143 | 0.314 | 0.487 | 0.599 | 0.440 | 0.322 |
| $\mathcal{L}_{RFCM}$ | 0.489 | 0.734 | 0.146 | 0.314 | 0.571 | 0.587 | 0.621 | 0.400 |
| $FCM$[18] | 0.061 | 0.987 | 0.012 | 0.031 | 0.089 | 0.983 | 0.046 | 0.047 |
| $RFCM$[28] | 0.288 | 0.871 | 0.089 | 0.168 | 0.714 | 0.834 | 0.666 | 0.555 |

TABLE III. Comparisons of computational time between the proposed method and other unsupervised methods.

Average Computational Time Per Image (sec)

| ConvNet with $\mathcal{L}_{MS}$ | ConvNet with $\mathcal{L}_{RFCM}$ | $FCM$ | $RFCM$ |
|---|---|---|---|
| 0.013 | 0.013 | 0.703 | 2.868 |

volumes were standardized to their Z-scores by subtracting their means and dividing by their standard deviations (i.e., $y_{z-score} = \frac{y - u_y}{\sigma_y}$). We validated the trained networks using the proposed loss function on 2 clinical SPECT scans and evaluated their performances on nine clinical scans. The segmentation performance of the proposed semi-supervised loss, $\mathcal{L}_{semi-RFCM}^{\alpha}$ using $q = 2$, was compared with that of a semi-supervised loss, Mumford-Shah loss (10),[37] and two supervised losses, Dice loss (17) and cross-entropy loss (18). Please note we also compared different values of $\alpha$, the weighting factor controlling the relative strength of the supervised and unsupervised loss functions. Figure 7 shows the segmentation results for four example slices of the clinical SPECT scan, from left to right: the first column displays the SPECT images; the second column shows the gold-standard segmentation provided by the radiologist, where the red contour indicates lesion and the green contour indicates bone; the third through last columns are the segmentation results generated by the ConvNet models trained using different loss functions. As indicated by the arrows in Fig. 7, the semi-supervised loss functions (the third through fourth columns) yielded better visual results than the supervised loss functions and the fixed-threshold method, where the semi-supervised models generally detected the locations and the shapes of the lesion and bone. The mean segmentation scores are shown in Table V, and statistical plots of the DSC and surface DSC scores are shown in the left panel of Fig. 6; the top two figures show the mean and standard deviation of the DSC scores, and the bottom figures present boxplots of the surface DSC. We find that the highest segmentation scores were also provided by the semi-supervised losses, in which the proposed $\mathcal{L}_{semi-RFCM}^{0.1}$ yielded 0.747 in DSC and 0.788 in surface DSC for lesion segmentation, and the highest bone segmentation scores with a DSC of 0.742 in DSC and 0.95 in surface DSC. Among the semi-supervised losses, the proposed $\mathcal{L}_{semi-RFCM}$ loss function outperformed the $\mathcal{L}_{semi-MS}$ loss function[37] in both lesion and bone segmentation tasks. When compared with the supervised segmentation loss functions, the proposed $\mathcal{L}_{semi-RFCM}$ outperformed them with $P$-values $< 0.001$ from a paired t-test. The supervised loss functions, $\mathcal{L}_{DSC}$, $\mathcal{L}_{CE}$, and the proposed $\mathcal{L}_{FCM_{label}}$, all failed in this case, producing unacceptable segmentation results with DSC less than 0.4 for lesion segmentation. This suggests that the simulation-trained ConvNets using the supervised losses did not generalize well on clinical SPECT data.

### 3.D. Semi-supervised Loss on Clinical CT

This experiment evaluates the performances of the proposed model applied to clinical CT images. In order to train a ConvNet to segment anatomical structures from CT images, a possible choice for the training dataset is the realistic CT simulations generated using attenuation maps of the XCAT phantom.[51,67] However, since the Hounsfield Unit (HU) in CT is simply a linear transformation of the linear attenuation coefficient of the human body, and the scope of this work is to explore the robustness of the proposed method in a setting where the images used for training and testing are from different domains, we instead, trained the ConvNets directly on the attenuation maps of a single XCAT phantom. We applied the

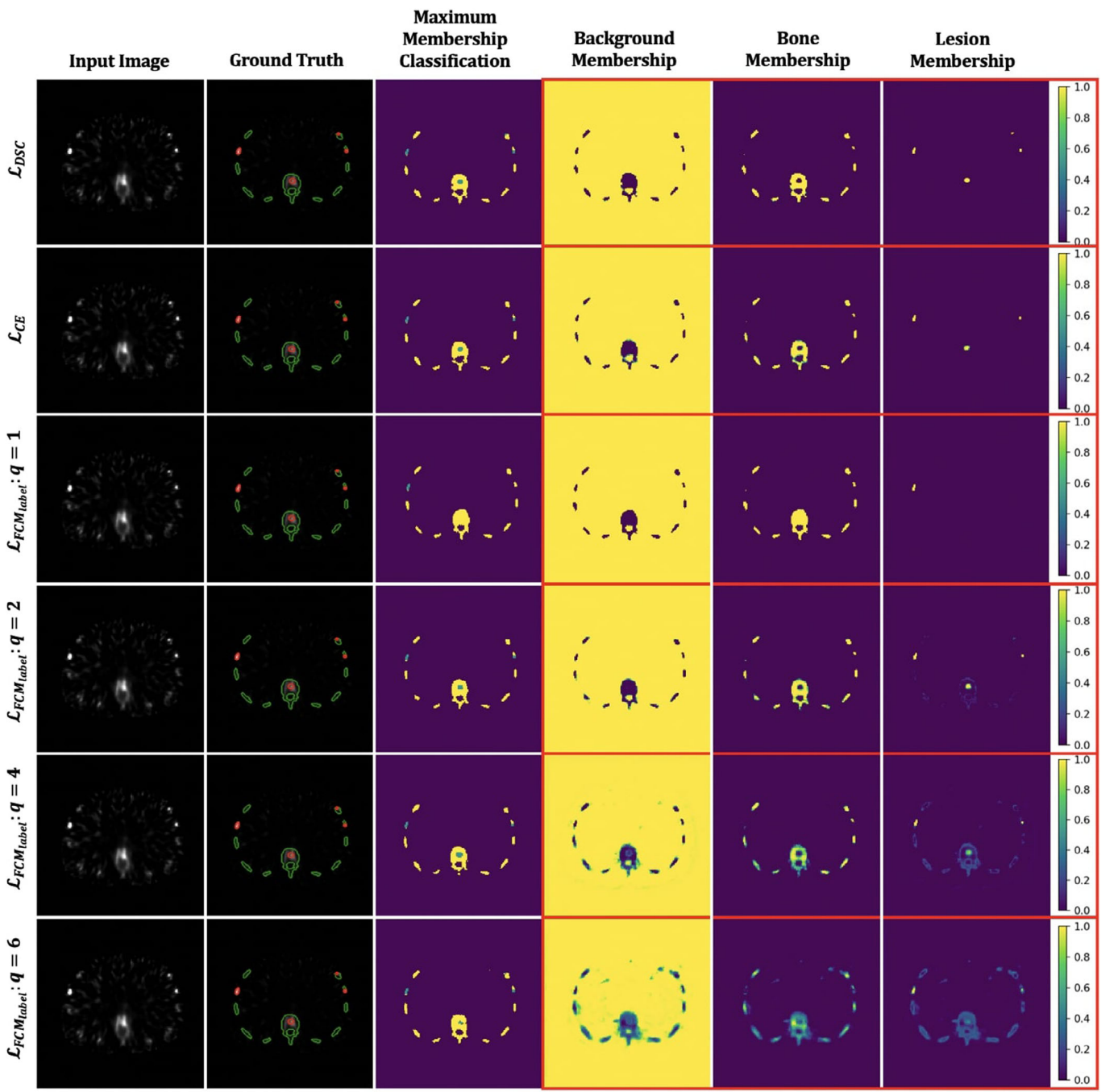

FIG. 5. Visual comparisons of the output membership functions generated using $\mathcal{L}_{DSC}$, $\mathcal{L}_{CE}$, the proposed $\mathcal{L}_{FCM_{label}}$ with different fuzzy exponent $q$. The red and green contours in the second column indicate truth regions for lesion and bone. The blue and yellow regions in the third column represent segmented lesion and bone.

same semi-supervised training scheme (with $q = 2$, and $\alpha = 0.3$ and $0.5$), as described in the previous section. The proposed method was applied to 12 clinical CT scans, where two scans were used for validation and 10 were used for evaluation. We normalized the 2D slices of the training and testing images so that the intensities of the slices had values in the interval $[0, 1]$. Data augmentation using Gamma correction was applied to the training images with $\gamma \in [1.4, 1.6, 1.8]$. The resulting segmentation was compared both qualitatively and quantitatively to the gold-standard segmentation provided by a radiologist. Figure 8 shows some qualitative segmentation results with the corresponding gold-standard segmentation. The proposed method achieved a more accurate bone delineation; specifically, the proposed method was able to detect the boundary of the bone region in more detail (as indicated by the yellow arrows). Table VI shows the mean segmentation scores, and the right panel of Fig. 6 shows statistical plots of the scores. The proposed method, $\mathcal{L}_{semi-RFCM}^{0.2}$, outperformed other loss functions by a significant margin with the high scores of 0.841 in surface DSC

TABLE IV.  Quantitative comparisons between the proposed supervised FCM loss, CE, and DSC loss on the realistic QBSPECT simulations.

| | Lesion | | | | Bone | | | |
|---|---|---|---|---|---|---|---|---|
| | DSC | Recall | Precision | IoU | DSC | Recall | Precision | IoU |
| $\mathcal{L}_{DSC}$ | 0.857 | 0.857 | 0.946 | 0.750 | 0.888 | 0.896 | 0.911 | 0.799 |
| $\mathcal{L}_{CE}$ | 0.858 | 0.957 | 0.838 | 0.751 | 0.898 | 0.894 | 0.937 | 0.815 |
| $\mathcal{L}_{FCM_{label}}^{q=1}$ | 0.559 | 0.493 | 0.941 | 0.388 | 0.901 | 0.907 | 0.928 | 0.820 |
| $\mathcal{L}_{FCM_{label}}^{q=2}$ | 0.859 | 0.910 | 0.894 | 0.753 | 0.906 | 0.915 | 0.930 | 0.828 |
| $\mathcal{L}_{FCM_{label}}^{q=4}$ | 0.851 | 0.883 | 0.917 | 0.741 | 0.901 | 0.940 | 0.899 | 0.820 |
| $\mathcal{L}_{FCM_{label}}^{q=6}$ | 0.852 | 0.919 | 0.886 | 0.742 | 0.900 | 0.908 | 0.927 | 0.818 |

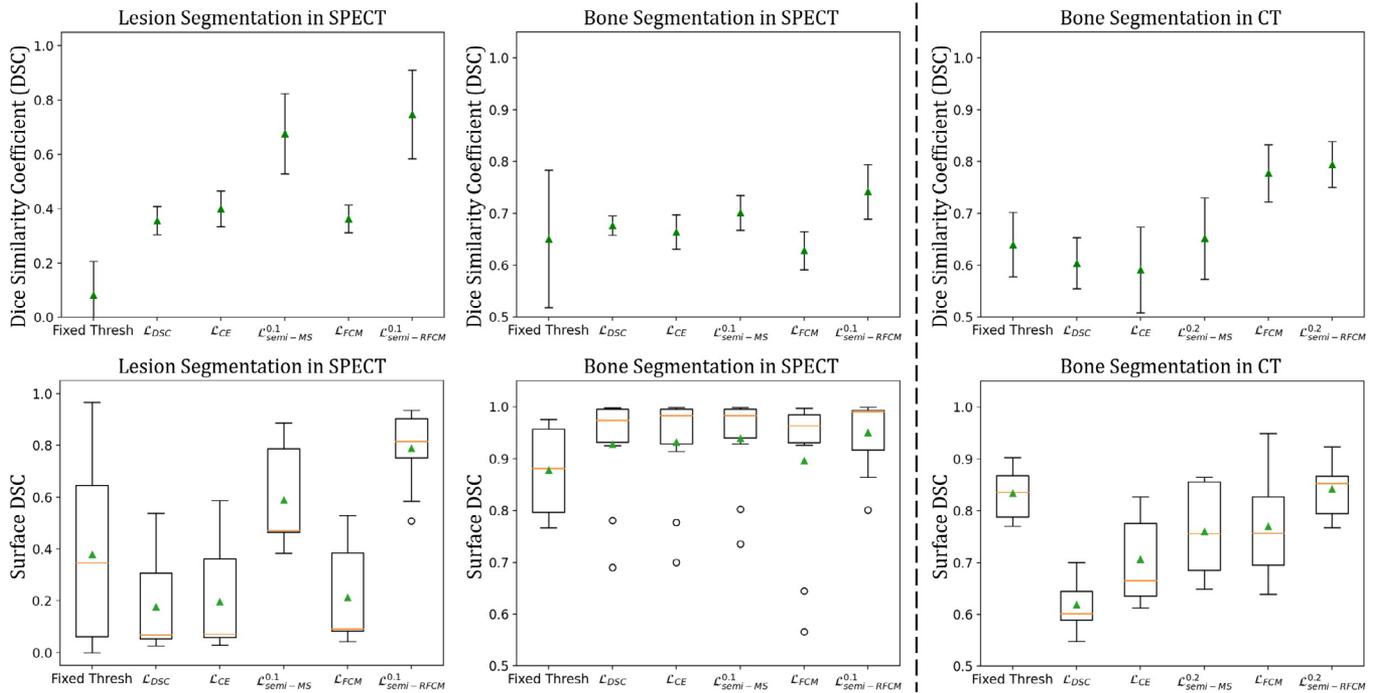

FIG. 6.  Statistical plots of DSC and surface DSC scores for SPECT and CT segmentation using different loss functions. The top three figures plot the mean scores in DSC with standard deviations as error bars. Figures in the bottom row are boxplots of the surface DSC, where the triangles denote the means, and the center lines represent the medians.

TABLE V.  Quantitative comparisons between the proposed semi-supervised and supervised losses, semi-supervised MS loss,[37] CE, and DSC loss on the dataset of clinical SPECT scans.

| | Lesion | | | | | Bone | | | | |
|---|---|---|---|---|---|---|---|---|---|---|
| | surface DSC | DSC | Recall | Precision | IoU | surface DSC | DSC | Recall | Precision | IoU |
| Fixed Threshold | 0.399 | 0.347 | 0.649 | 0.197 | 0.124 | 0.877 | 0.651 | 0.734 | 0.716 | 0.496 |
| $\mathcal{L}_{DSC}$ | 0.177 | 0.356 | 0.297 | 0.473 | 0.218 | 0.928 | 0.677 | **0.810** | 0.589 | 0.512 |
| $\mathcal{L}_{CE}$ | 0.197 | 0.400 | 0.348 | 0.498 | 0.252 | 0.931 | 0.664 | 0.737 | 0.614 | 0.498 |
| $\mathcal{L}_{semi-MS}^{0.1}$ | 0.590 | 0.677 | 0.650 | 0.709 | 0.527 | 0.939 | 0.701 | 0.798 | 0.637 | 0.541 |
| $\mathcal{L}_{FCM_{label}}$ | 0.213 | 0.363 | 0.350 | 0.433 | 0.223 | 0.896 | 0.628 | 0.784 | 0.550 | 0.459 |
| $\mathcal{L}_{semi-RFCM}^{0.1}$ | **0.788** | **0.747** | **0.725** | **0.774** | **0.618** | **0.950** | **0.742** | 0.725 | **0.774** | **0.592** |

The top performances are shown in bold.

and 0.794 in DSC. Interestingly, the proposed supervised loss, $\mathcal{L}_{FCM_{label}}$, achieved 0.770 in surface DSC and 0.778 in DSC, which were the highest among the supervised loss functions. The proposed semi-supervised loss function yielded $p-$values $<0.001$ from the paired t-test was obtained when it was compared to $\mathcal{L}_{DSC}$, $\mathcal{L}_{CE}$, and $\mathcal{L}_{semi-MS}^{0.2}$.

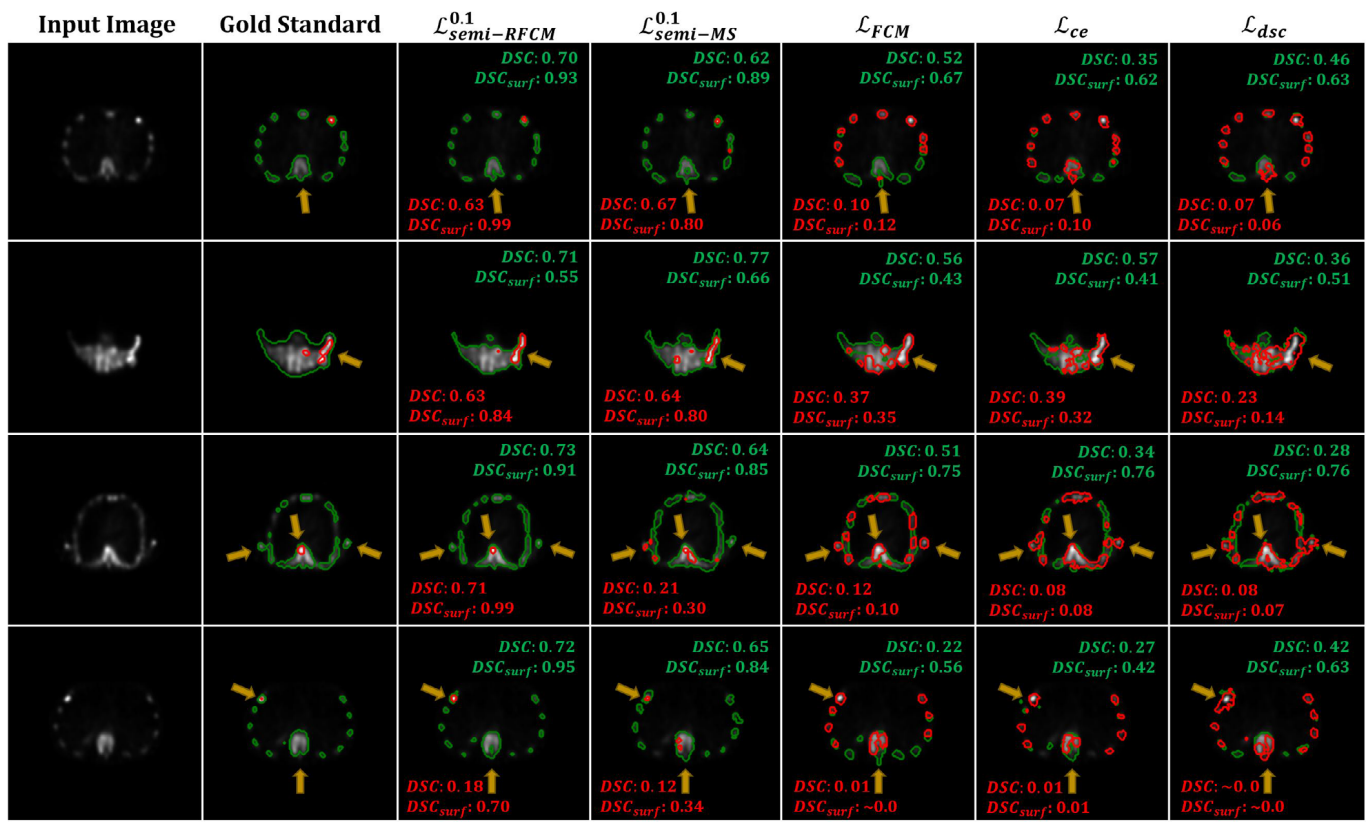

FIG. 7.  Qualitative comparison results of the proposed semi-supervised loss and other losses on the clinical SPECT scan. The yellow arrows highlight the differences between the segmentation.

## 4.  DISCUSSION

### 4.A.  Unsupervised model

In the traditional paradigm of learning-based methods, there often exists a prior training stage that involves ground truth labels. In this study, we presented, by contrast, a novel unsupervised loss function that extends the ability of neural networks to perform unsupervised segmentation without the need for ground truth labels at any stage. The quantitative results shown in Table I and II indicate that, although the unsupervised loss functions yielded a low DSC that was around 0.50, they provided recalls of $>72\%$. In $^{99m}$ Tc-MDP-based bone SPECT, lesions often have the highest mean uptake, followed by bone and soft tissue. The proposed unsupervised ConvNets were trained to distinguish different characteristics of intensity distributions between different organs. As a result, the proposed model was often able to capture lesion and bone as two separate classes, thus leads to higher recall. We applied a network visualization method proposed by Erhan et al.[68] to visualize some example filters in the ConvNets. Erhan's method maximizes a particular filter's activation by updating a random image. We refer the readers to[68] for the implementation details. The visualizations of five filters from the second-to-last layer of the ConvNets are shown in Fig. 9, where the first row exhibits the visualizations for the proposed unsupervised ConvNet, and the second through the last row show the visualizations for the supervised models. These images demonstrate that the unsupervised

ConvNet produced filter patterns, representing image features, that are drastically different from those of the other methods. The supervised ConvNets performed well at differentiating shapes and patterns in images.[68,69] Therefore, the differences observed in filter visualizations imply that the proposed unsupervised training scheme considered additional information other than shapes and patterns within a given image.

We also demonstrated that the proposed unsupervised ConvNet model was robust to noise in the test image. It is important to mention that the ConvNets were trained on a noise-free dataset, but the ConvNet architecture and the MRF-based regularizer (4) in $\mathcal{L}_{RFCM}$ were able to provide good noise suppressing performances that outperformed the TV regularizer used in $\mathcal{L}_{MS}$.[37] This improvement was also demonstrated by the results shown in Table II and Fig. 4. Specifically, $\mathcal{L}_{RFCM}$ outperformed $\mathcal{L}_{MS}$ by 0.01 in DSC for lesion segmentation and 0.08 in DSC for bone segmentation on a Gaussian noise corrupted test image set.

The proposed $\mathcal{L}_{RFCM}$ loss function incorporates an important idea from FCM, where there is a parameter, $q$, that allows for the non-binary membership functions. The effects of different values of $q$ are shown in Fig. 4, where smaller $q$ values lead to crisper membership values. As described previously, the formulation of $\mathcal{L}_{MS}$ in Ref. [37] is similar to a special case of $\mathcal{L}_{RFCM}$ where the fuzzy exponent, $q$, is 1. Therefore, the membership functions produced by training with $\mathcal{L}_{MS}$ resulted in "hard" (non-fuzzy) membership values, which

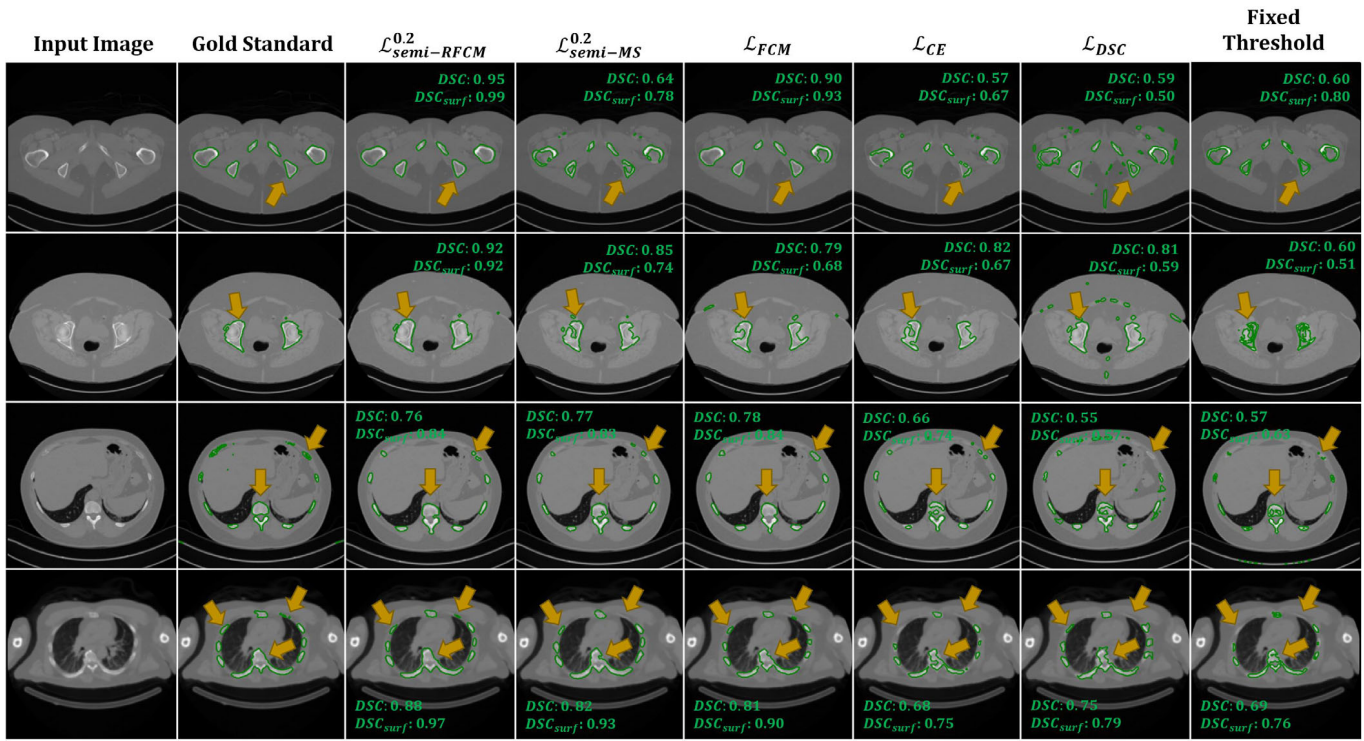

FIG. 8. Qualitative comparison of bone segmentation results between gold-standard and ConvNets trained using different losses on the clinical CT. The yellow arrows highlight the differences between the segmentation.

TABLE VI. Quantitative comparisons between the proposed semi-supervised RFCM loss, semi-supervised MS loss, CE, and DSC loss on the clinical CT scan.

|  | Surface DSC | DSC | Recall | Precision | IoU |
|---|---|---|---|---|---|
| Fixed Threshold | 0.834 | 0.640 | 0.492 | **0.933** | 0.473 |
| $\mathcal{L}_{DSC}$ | 0.614 | 0.604 | 0.703 | 0.548 | 0.436 |
| $\mathcal{L}_{CE}$ | 0.707 | 0.591 | 0.462 | 0.848 | 0.425 |
| $\mathcal{L}_{semi-MS}^{0.2}$ [37] | 0.760 | 0.652 | 0.552 | 0.812 | 0.489 |
| $\mathcal{L}_{FCM_{label}}$ | 0.770 | 0.778 | 0.877 | 0.708 | 0.640 |
| $\mathcal{L}_{semi-RFCM}^{0.2}$ | **0.841** | **0.794** | **0.790** | 0.804 | **0.661** |

The top performances are shown in bold.

were similar to those produced by training using $\mathcal{L}_{RFCM}$ with $q = 1$.

## 4.B.  Supervised model

The proposed supervised loss function was inspired by,[49] for which Chen et al. modified the objective function of the well-known Active contour without edges (ACWE) to operate on the ground truth segmentation during training the ConvNet. With a similar idea, we proposed a novel supervised loss function that is based on the objective function of the classical FCM. This loss function, $\mathcal{L}_{FCM_{label}}$, shares the property of $\mathcal{L}_{RFCM}$ that it incorporates a parameter that supports fuzziness in the membership function. This enabled us to combine $\mathcal{L}_{FCM_{label}}$ with $\mathcal{L}_{RFCM}$ as a new loss function that can be minimized by a ConvNet in a semi-supervised manner. To the best of our knowledge, this is the first work that introduces the idea of fuzzy clustering into the loss function of CNN-based segmentation.

The effectiveness of $q$ in $\mathcal{L}_{FCM_{label}}$ is visually shown in Fig. 5. As opposed to the membership values generated by $\mathcal{L}_{CE}$ and $\mathcal{L}_{DSC}$ which were crisper, $\mathcal{L}_{FCM_{label}}$ offered different degrees of fuzzy overlap between resulting membership functions by selecting different values of $q$. Unlike hard membership functions, the fuzzy exponent, $q$, enables a voxel to be in multiple classes. This could be useful in quantifying nuclear medicine images as the images often exhibit partial volume effects, where a voxel usually contains uptake information from more than one class. Quantitatively, on a test dataset comprised of simulated QBSPECT images, a mean DSC of 0.85 produced by the ConvNets trained using $\mathcal{L}_{FCM_{label}}$ was comparable to those trained using $\mathcal{L}_{CE}$ and $\mathcal{L}_{DSC}$ (as shown in Table IV). Interestingly, although the ConvNet models trained using the supervised losses did not perform well on the clinical dataset, the proposed $\mathcal{L}_{FCM_{label}}$ still outperformed the other two supervised losses, particularly by 0.035 in DSC on the bone segmentation from the clinical CT.

## 4.C.  Semi-supervised Model

We combined the proposed unsupervised and supervised losses in a novel loss function, $\mathcal{L}_{semi-RFCM}^{\alpha}$, that enables the ConvNet to perform semi-supervised segmentation. This loss function was evaluated on a clinical SPECT/CT scan. We emphasize that the ConvNet models were purely trained using QBSPECT simulations and attenuation maps from the XCAT phantom. As shown in Tables V and VI, in general, semi-supervised training outperformed supervised training by a large margin of ~0.4 in DSC and surface DSC for lesion segmentation on the clinical SPECT and ~0.08 in DSC on the clinical CT. This is because of the poor generalizability provided by the ConvNets when they are trained using supervised loss functions on purely simulation images, for which they tend to over-fit the semantic information in the training data. Therefore, the performance degradation is expected when testing (i.e., clinical studies) and training data (i.e., simulated images) are not drawn from the same distribution. In comparison, the semi-supervised losses impose the ConvNet models to consider both pixel-level intensity distributions, provided by minimizing the unsupervised losses, and semantic information, provided by minimizing the supervised losses, of a given image. Notice that, as highlighted by the arrows Fig. 10, there are some clear errors in the "gold-standard" segmentation. In comparison, the proposed model would seem to have provided a more accurate delineation. The regions with higher intensity levels and the regions within cortical bones (i.e., bone marrow) were all correctly being classified as bone by the automated observers, in contrast to the human observers. The parameter $\alpha$ is a hyper-parameter that controls the weight of the supervised loss.

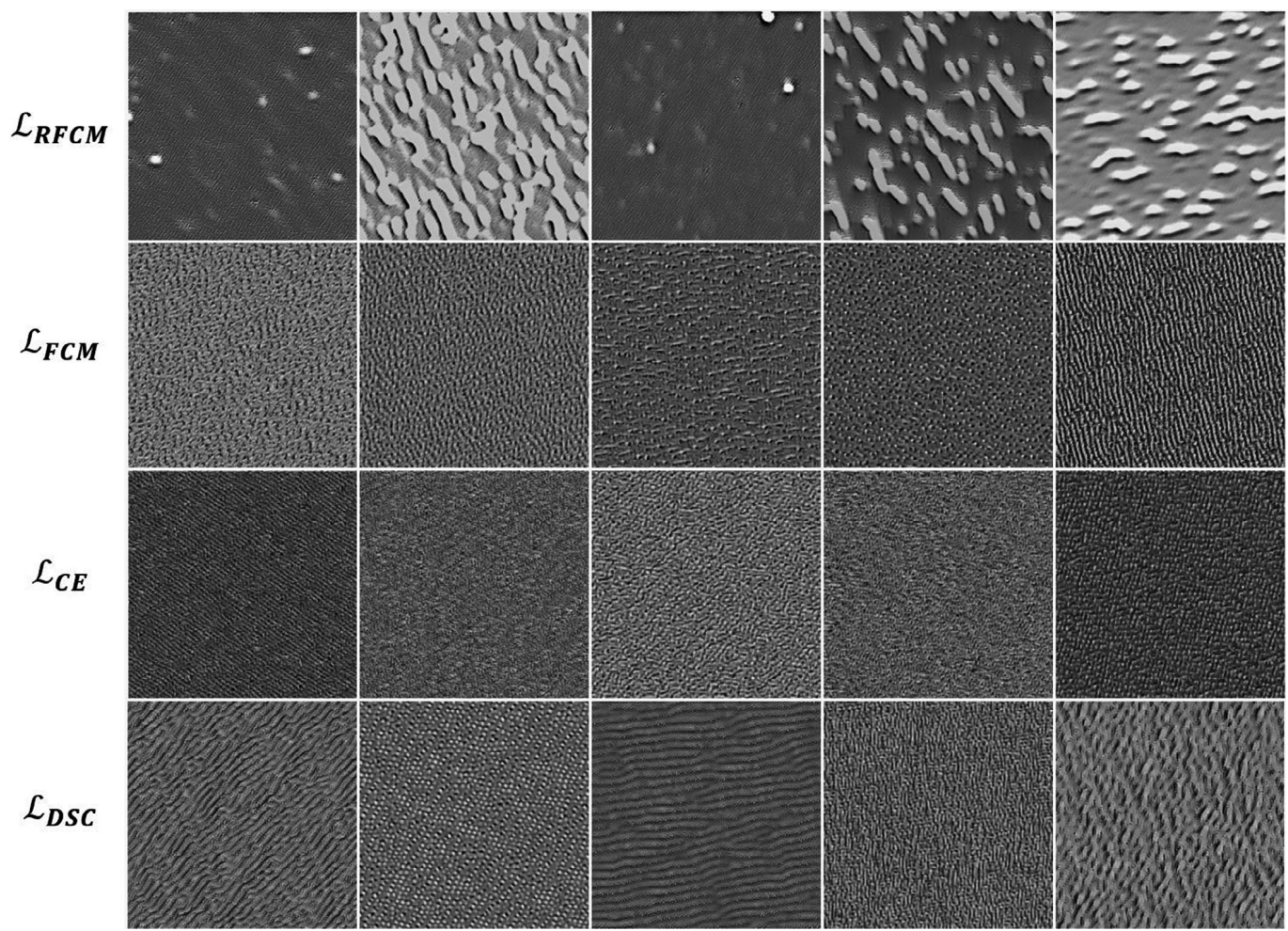

FIG. 9. Visualization of five filters from the second-to-last convolutional layer of the networks trained using un- and fully-supervised loss functions.

When $\alpha$ is set to be small, the network focuses more on characterizing intensity distributions rather than the segmentation labels from the training dataset. The impact of $\alpha$ in $\mathcal{L}^{\alpha}_{semi-RFCM}$ ($q=2$) and $\mathcal{L}^{\alpha}_{semi-MS}$ on the performance of segmenting the clinical data are quantitatively studied, and the results are shown in Fig. 11. For the SPECT data, $\mathcal{L}^{\alpha}_{semi-RFCM}$ outperforms $\mathcal{L}^{\alpha}_{semi-MS}$ in lesion segmentation for all the $\alpha$ values tested, while sharing the similar performances in bone segmentation. For the CT data, the performance of $\mathcal{L}^{\alpha}_{semi-RFCM}$ was consistently higher than that of $\mathcal{L}^{\alpha}_{semi-MS}$ for all $\alpha$ values we tested.

One focus of this work has been to present a series of loss functions to use ConvNets that require limited amounts of training data. The work concentrated on the development of the loss functions, and an effort to optimize the ConvNet architecture is an interesting avenue for future study. A limitation of this work is that the amount of clinical data used for evaluation is relatively small. However, the existing results have demonstrated that the proposed semi-supervised loss function successfully enabled ConvNets trained using purely simulated images that were totally independent of the clinical images to produce usable segmentation for clinical images. Further evaluation using

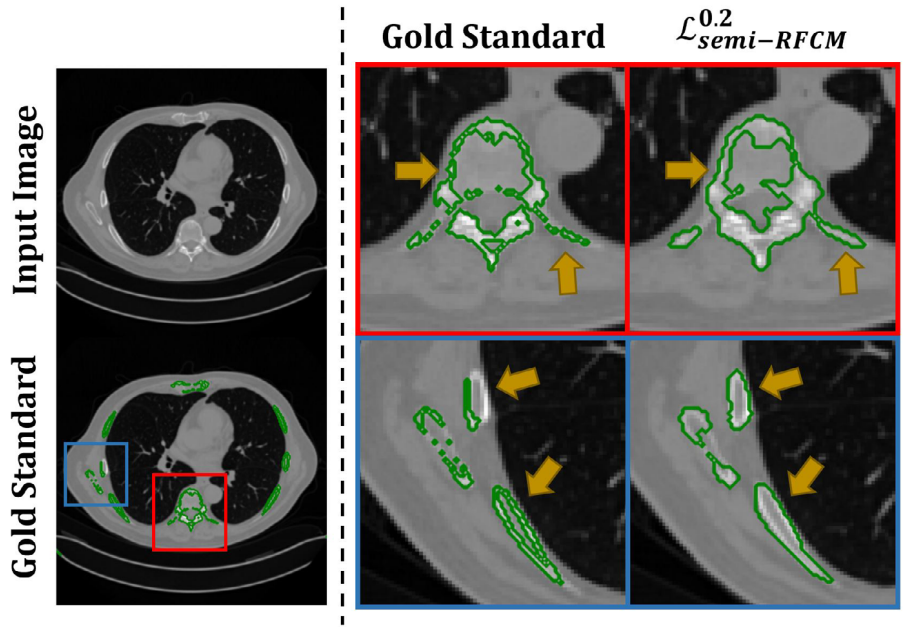

FIG. 10.  Left panel: A slice of CT image (top). The "gold-standard" delineation (bottom). Right panel: Magnified segmentation results. The first column denotes the "gold-standard" segmentation, and the second column denotes the results obtained using the proposed method.

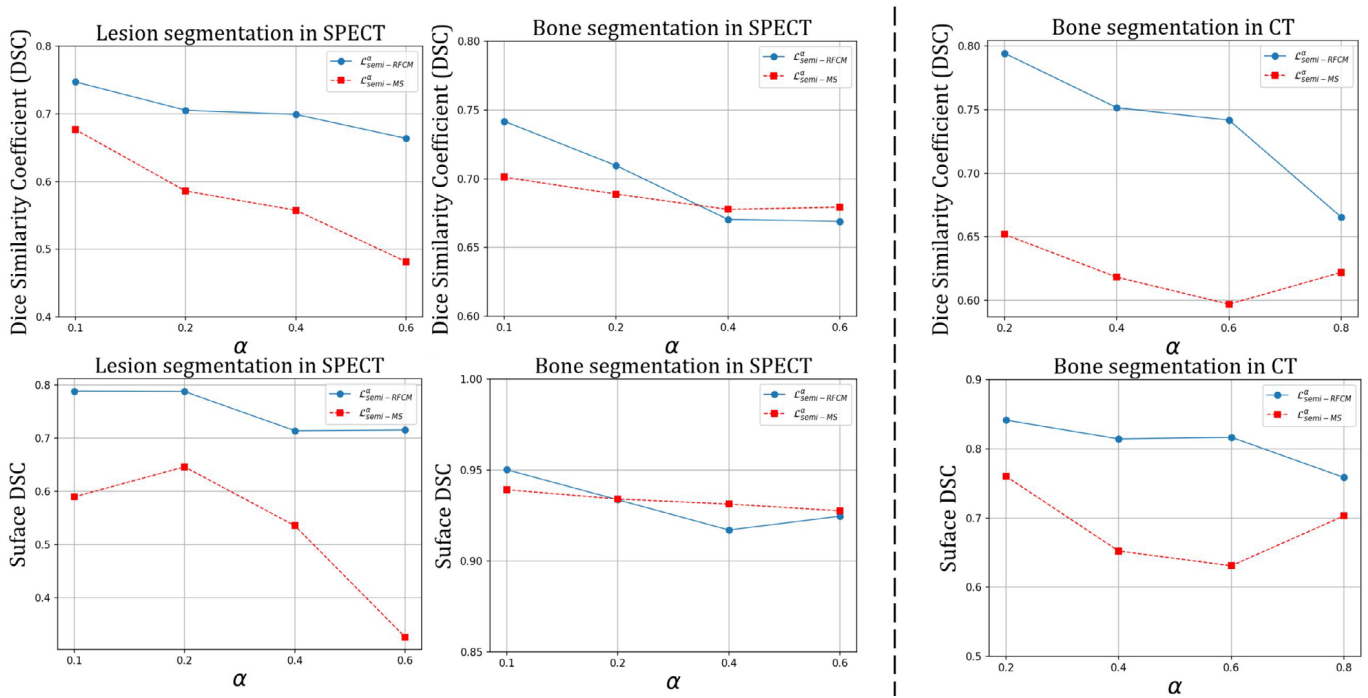

FIG. 11.  The impact of different $\alpha$ values in the semi-supervised losses on the performance of segmenting clinical SPECT (a), and CT (b).

more clinical images will be needed before the clinical application of the proposed method.

## 5.  CONCLUSION

In this work, we proposed a set of novel FCM-based loss functions for semi-, unsupervised, and supervised SPECT/ CT segmentation using deep neural networks. An advantage of the proposed loss functions is that they enable the ConvNets to consider both voxel intensity and semantic information in the image during the training stage. The proposed loss functions also retain the fundamental property of the conventional fuzzy clustering, where the fuzzy overlap between the channels of softmax outputs can be adjusted by a hyper-parameter in the loss function. Various experiments demonstrated that the model trained using a dataset of simulated images generalized well and led to fast and robust segmentation on both simulated and clinical SPECT/CT images.

## ACKNOWLEDGMENTS

This work was supported by a grant from the National Cancer Institute, U01-CA140204. The views expressed in written conference materials or publications and by speakers and moderators do not necessarily reflect the official policies of the NIH; nor does mention by trade names, commercial practices, or organizations imply endorsement by the U.S. Government.

## CONFLICT OF INTEREST

The authors have no conflicts to disclose.

### Data Availability Statement

The data used in this study were acquired as part of an IRB protocol and is not approved for public release.

a)Author to whom correspondence should be addressed. Electronic mail: jchen245@jhmi.edu.